\documentclass{article}

\usepackage[dblblindworkshop,final]{neurips_2026}
\workshoptitle{The 6th Workshop on Mathematical Reasoning and AI}

\usepackage[utf8]{inputenc}
\usepackage[T1]{fontenc}
\usepackage{hyperref}
\usepackage{url}
\usepackage{booktabs}
\usepackage{amsmath}
\usepackage{amsfonts}
\usepackage{amssymb}
\usepackage{graphicx}
\usepackage{subcaption}
\usepackage{nicefrac}
\usepackage{microtype}
\usepackage{xcolor}
\usepackage{float}

\title{Auditing LLM Program Search for Bermudan\\Stopping with Primal--Dual Certificates}

\author{%
  Yuhua Qian\\
  Shanghai Artificial Intelligence Laboratory\\
  \texttt{qianyuhua@pjlab.org.cn}
}

\begin{document}

\maketitle

\begin{abstract}
LLM program-search agents are usually judged by the score that drives their search.
We study LLM-driven search over Python exercise policies for Bermudan options, an optimal stopping problem where a trusted evaluator owns the simulation and a dual upper bound~$U$.
For any policy, $0 \le V^\star - L_\pi \le U - L_\pi$; we report Monte Carlo estimates of both bounds with their sampling error.
We audit 24 generation attempts from two pipelines, a six-pair feedback-representation ablation, and a holdout evaluated after selection was frozen.
No selected generated program improves on Ridge LSMC consistently or by more than Monte Carlo uncertainty; holdout gap scores lie within $0.002$ of it.
On six one-dimensional puts, a high-precision binomial reference places Ridge within Monte Carlo noise of optimal and attributes 99\% of the estimated gap to dual-bound looseness, so this dual audits validity but cannot certify progress at the search's scale.
Structured memory showed no benefit over score-only memory in this ablation.
Whether search ``worked'' depends on reporting rules: retaining the initial reference, freezing selection before holdout, and counting failures.
\end{abstract}

\section{Introduction}
\label{sec:intro}

LLM program-search systems have produced mathematical results \citep{romera2024mathematical,novikov2025alphaevolve}, and agentic harnesses that propose their own experiments are under development \citep{lehman2022elm,lu2024aiscientist,hyra2026}. In each, the objective that drives search is also the reported result. When that objective is a noisy Monte Carlo estimate, the proxy can improve while the underlying quantity stays flat \citep{manheim2018goodhart}, and a reader cannot separate the two cases.

Where the search objective is an exact verifier, the score is already a certificate of quality. Where it is a Monte Carlo estimate, a certificate must come from the task's own duality. Bermudan option pricing has one. A Bermudan option gives its holder the right to collect a payoff, such as $\max(K - S, 0)$ for a put on an asset with price $S$, at any one of a finite set of dates. Its price is the value of an optimal stopping problem, so a suboptimal exercise policy misprices the contract. With several risk factors, where lattice methods are impractical, contracts such as callable bonds are priced by simulation with regression-based exercise policies \citep{longstaff2001valuing,glasserman2004monte}. Three properties make the problem a clean test of program search. A policy is a short program that maps the current state to exercise or continue. Its value can only be estimated by noisy simulation, the regime in which search scores mislead. And duality gives an upper bound on the optimal value \citep{rogers2002monte,haugh2004pricing,andersen2004primal}, so the distance of a candidate from optimal can be bounded without knowing the optimum, by trusted code the candidate cannot touch. Deep learning methods for optimal stopping exist \citep{becker2019deep}; we study the evaluation, not the algorithm.

We build an open-source planner$\rightarrow$proposer search harness (\url{https://github.com/MrSteeeve/OpenHyra}), run it on Bermudan optimal stopping, and re-evaluate every candidate with primal--dual gap certificates. Our contributions:
\begin{enumerate}\setlength\itemsep{0pt}
\item A certificate-based evaluation protocol (\S\ref{sec:cert}--\S\ref{sec:harness}): a trusted evaluator owns the simulation, causal exercise, and a shared dual bound; candidates control only when to stop. We state what the Monte Carlo estimate of the gap does and does not guarantee.
\item An audit of 24 generation attempts, a feedback-representation ablation, and a holdout evaluated after selections were frozen. No selected generated program improves on Ridge least-squares Monte Carlo (LSMC) consistently or by more than Monte Carlo uncertainty; all gap scores lie within $\pm 0.002$ of it. The one selection-score gain is an artifact of the score and did not transfer.
\item A calibration against high-precision reference values showing that the dual bound, not the policy, dominates the estimated gap, and that the reference puts leave little room for any policy to improve.
\item A reporting-rule analysis showing that implicit conventions determine whether the search appears to have worked.
\end{enumerate}

\section{Problem and certificate}
\label{sec:cert}

A Bermudan contract may be exercised at dates $m=0,\dots,M$. Let $Z_m = e^{-rt_m}\,\mathrm{payoff}(S_m)$ be the discounted payoff and $V^\star = \sup_{\tau \in \mathcal{T}} \mathbb{E}[Z_\tau]$ the target \citep{glasserman2004monte}. A candidate policy $\pi$ at date $m$ sees only $S_m$, the time index, and public parameters; the evaluator enforces causality, allows at most one exercise, and forces settlement at maturity. With $L_\pi = \mathbb{E}[Z_{\tau_\pi}]$, for any valid dual upper bound $U$:
\begin{equation}
L_\pi \;\le\; V^\star \;\le\; U,
\qquad\text{so}\qquad
0 \;\le\; V^\star - L_\pi \;\le\; U - L_\pi\,.
\label{eq:cert}
\end{equation}
For the true values $L_\pi$ and $U$, the gap $U - L_\pi$ certifies suboptimality of $\pi$ without requiring $V^\star$. All candidates in a single evaluation request share the same evaluator-owned $U$, so under that common $U$ a smaller gap is equivalent to a higher policy value. Comparing two policies is therefore a paired payoff comparison in which $U$ cancels; the gap adds information only about distance from optimal.

The dual uses an evaluator-owned Ridge-LSMC value proxy $f_m$ \citep{longstaff2001valuing,tsitsiklis2001regression} in a nested martingale: $M_0 = 0$, $\Delta M_m = f_m(S^{\mathrm{out}}_m) - \frac{1}{B}\sum_b f_m(S^{\mathrm{in},b}_m \mid S^{\mathrm{out}}_{m-1})$. The increments are conditionally mean-zero, making $\mathbb{E}[\max_m(Z_m - M_m)]$ a valid upper bound \citep{rogers2002monte,haugh2004pricing}. The nested budget is finite; $U$ is loose on some contracts, and candidates cannot tighten the shared bound.

\paragraph{What the estimates guarantee.} We observe Monte Carlo estimates $\hat L$ and $\hat U$ on independent paths, so a reported $\hat U - \hat L$ is an estimate of the certificate width, not itself a certificate. $\hat L$ is computed on pricing paths independent of the paths used to fit the policy, so it is unbiased for $L_\pi$; the inner averages make $M$ a martingale, so $\hat U$ is unbiased for a valid upper bound. By the central limit theorem, $\hat U_i + z\,\mathrm{SE}_{U,i}$ and $\hat L_i - z\,\mathrm{SE}_{L,i}$ are approximate one-sided confidence bounds on $U_i$ and $L_{\pi,i}$, and by Bonferroni over the two endpoints
\[
g_i = \bigl(\hat U_i - \hat L_i + 1.96\,(\mathrm{SE}_{U,i} + \mathrm{SE}_{L,i})\bigr)/K_i
\]
is an approximate one-sided 95\% upper confidence bound on $(V^\star_i - L_{\pi,i})/K_i$, where $K_i$ is the contract's strike or notional. Simultaneous coverage over $n$ instances needs $z = \Phi^{-1}(1 - 0.05/2n)$, e.g.\ $z = 2.50$ for the four holdout contracts of a seed. These statements are asymptotic; we give no finite-sample guarantee, and $\hat U$ is heavy-tailed on some contracts (Appendix~\ref{app:decomp}).

\paragraph{Gap score.} The raw gap is $(\hat U_i - \hat L_i)/K_i$. As a selection diagnostic we also compute $S = \mathrm{mean}_i(g_i) + 0.25\,Q_{90}(g)$, where $Q_{90}$ is the 90th percentile and $0.25$ is a legacy heuristic. Because $S$ adds the sampling margin to the raw gap, it rewards smaller sampling error as well as higher payoff and can move without the raw gap moving (\S\ref{sec:results}). We report $\hat L$, the raw gap $(\hat U{-}\hat L)/K$, and $S$ separately.

\section{Harness and protocol}
\label{sec:harness}

\begin{figure}[t]
\centering
\includegraphics[width=\textwidth]{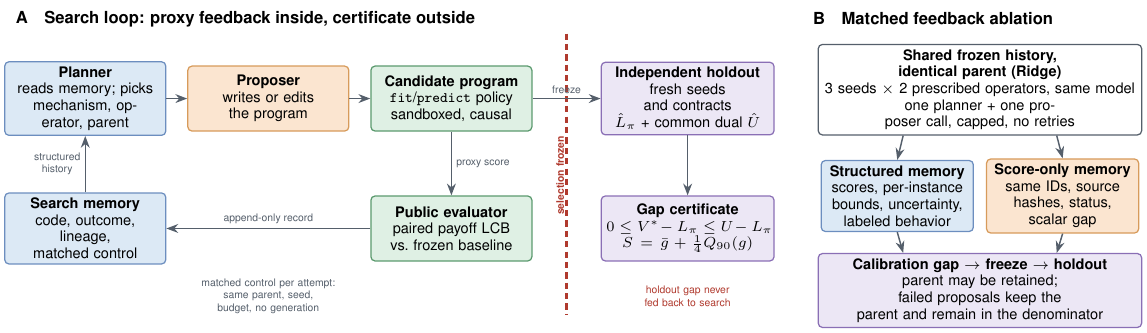}
\caption{Executed pipeline. A public payoff-improvement LCB drives search; selections freeze before the independent holdout evaluation, and the holdout gap is never fed back (the ablation uses the calibration score $S$ for feedback and selection). The dual bound is evaluator-owned and shared across all candidate Python policies.}
\label{fig:pipeline}
\end{figure}

\paragraph{Pipelines.}
Two pipelines were run with real model calls on three seeds (model and backend in Appendix~\ref{app:protocol}). (a)~Planner$\rightarrow$proposer: a planner reads a structured search memory and picks a mechanism, operator, and parent; a proposer then writes or edits the program. No weights are updated; adaptation is through prompts and parent choice. (b)~Direct generation: no planner, frozen reference parent. Each pipeline ran $3 \times 2 \times 2 = 12$ attempts with operators prescribed per round (round~0: whole-program rewrite or AST-subtree mutation; round~1: AST crossover or single-component rewrite), and every attempt had a matched no-generation control. ``Public'' feedback (visible to the agent during search) was a paired payoff-improvement 95\% lower confidence bound (LCB), not the gap. Failures: $2/12$ and $3/12$ proposer timeouts, kept in all denominators. The two pipelines differ in call budget and parent adaptivity; the comparison is structural, not equal-compute.

\paragraph{Data tiers.}
The incumbent is the currently selected program, which may remain the initial reference (Ridge). Evaluation uses four tiers with a hard freeze between selection and holdout: (i)~public search; (ii)~a calibration panel used for feedback and selection, plus $7$ reference programs evaluated retrospectively (two of which fit the same regression; Appendix~\ref{app:protocol}); (iii)~selection freeze; (iv)~an independent holdout on fresh seeds and contracts (puts, multi-asset max-calls, basket puts) over 37 programs. Path budgets grow across tiers (Appendix~\ref{app:protocol}).

\paragraph{Feedback ablation.}
Does the planner do better reading structured memory than scalar history? We fix history and parent and vary only the memory given to the planner: structured (scores, per-instance bounds, uncertainty, labeled behavior records) versus score-only (candidate IDs, source hashes, status, scalar gaps). The parent is Ridge. Model, operators, call caps, retry policy, and output schema are matched. Both arms use $S$ for feedback and selection, and the parent may be retained. This gives 6 matched pairs across 3 independent seed blocks. An infrastructure event was recovered before any proposer call (Appendix~\ref{app:recovery}).

\section{Results}
\label{sec:results}

\begin{table}[t]
\caption{Independent holdout, mean over three seeds (lower is better). $\hat U/K = 0.25537$ is shared by every row, so raw-gap differences ($(\hat U{-}\hat L)/K$) are payoff differences with the opposite sign. $S$ is the selection diagnostic (\S\ref{sec:cert}); SD is the descriptive spread across the three seed blocks (contracts differ by seed). Rows sharing source code produce correlated evaluations. $^\dagger$The PCA + Ridge and policy-iteration programs fit the same regression and give identical results (Appendix~\ref{app:protocol}).}
\label{tab:main}
\centering
\footnotesize
\begin{tabular}{lrrr}
\toprule
Method or selection rule & $\hat L/K$ & $(\hat U{-}\hat L)/K$ & Gap score (SD) \\
\midrule
Ridge LSMC & 0.15105 & 0.10433 & 0.18527 (0.09656) \\
PCA + Ridge / Policy iteration$^\dagger$ & 0.15124 & 0.10413 & \textbf{0.18504} (0.09621) \\
Gated Ridge & 0.15034 & 0.10503 & 0.18601 (0.09597) \\
MLP & 0.14170 & 0.11367 & 0.19560 (0.09292) \\
Ridge + MLP residual & 0.15075 & 0.10462 & 0.18570 (0.09567) \\
Threshold policy & 0.05858 & 0.19679 & 0.28960 (0.07331) \\
\midrule
Original planner: best generated & 0.14988 & 0.10550 & 0.18656 (0.09578) \\
Original direct: best generated & 0.15014 & 0.10524 & 0.18637 (0.09549) \\
Original planner incumbent & 0.15054 & 0.10483 & 0.18590 (0.09517) \\
Original direct incumbent & 0.15041 & 0.10496 & 0.18604 (0.09606) \\
Structured-memory incumbent & 0.15105 & 0.10433 & 0.18527 (0.09656) \\
Score-only incumbent & 0.15068 & 0.10469 & 0.18565 (0.09590) \\
\bottomrule
\end{tabular}
\end{table}

\paragraph{F1. No selected generated program improves on Ridge.}
Table~\ref{tab:main} shows that every LSMC-family reference and every generated program falls in the gap-score range $0.18504$--$0.18656$, a span of $0.0015$, below the per-contract standard error of $\hat L/K$ (about $0.003$ on the puts, Appendix~\ref{app:tightness}). The only large deviations are a plain MLP proxy ($0.19560$) and a hand-written threshold policy ($0.28960$, whose low $\hat L = 0.05858$ means its gap is lost payoff, not a loose bound). The best generated program from either pipeline is worse than its Ridge parent on average (Fig.~\ref{fig:results} in Appendix~\ref{app:figures}); one unselected candidate scores below Ridge on one seed (F4).

\paragraph{F2. A proxy gain that did not transfer.}
All six public winners in the original search had payoff LCB ${<}\,0$; per pipeline $3/12$ predictions pointed in the right direction with $0/12$ having confidence intervals above zero. In the ablation, one attempt improved its selection score: a score-only component rewrite at seed~17, with calibration $S = 0.200843$ against Ridge's $0.201146$. On holdout it was worse ($0.075803$ vs.\ $0.074647$). The decomposition (Appendix~\ref{app:decomp}) shows it had a lower payoff estimate ($\hat L = 0.192467$ vs.\ $0.192648$) and a larger raw gap ($0.111309$ vs.\ $0.111128$); $S$ fell only because the Monte Carlo margin shrank. The gain is an artifact of a score that rewards smaller sampling error: a paired payoff comparison (where $U$ cancels) would not have selected this candidate, since its payoff estimate is below the parent's.

\paragraph{F3. Structured memory: no benefit observed in this ablation.}
All four valid structured proposals and five of six score-only proposals were worse than their parent on calibration; the sixth is the score-only rewrite of F2 ($+0.000303$), which did not transfer (per-pair values in Appendix~\ref{app:tables}). On holdout the structured incumbent changed by $0$ in all three seeds (it retained Ridge every time); the score-only incumbent changed by $-0.00116/0/0$, at comparable model time ($773.9$ vs.\ $812.4$\,s). Structured memory produced $4/6$ valid attempts against $6/6$ for score-only; both timeouts occurred on whole-program restart, the operator with the largest prompt. Two timeouts in six pairs are limited evidence of a systematic difference; prompt length and content are confounded. A score-only summary of the ablation would hide the timeouts entirely.

\paragraph{F4. Reporting rules change the verdict.}
Whether this search ``worked'' depends on conventions that are usually left unstated. If the initial reference counts as a deployable incumbent, both original pipelines finish at $0.18590$/$0.18604$, better than their best generated program; the search helped only by not being followed. Dropping failures from denominators would hide the reliability gap between the ablation arms. With access to holdout numbers, we could report an unselected structured candidate at seed~17 that beats Ridge on holdout ($0.074467$ vs.\ $0.074647$). Selection was frozen before the holdout, so the holdout candidate is reported as an observation.

\paragraph{F5. Certificate tightness against reference values.}
We computed high-precision reference values $V^\star$ for the six one-dimensional holdout puts with a Cox--Ross--Rubinstein binomial tree, exercise restricted to the evaluator's dates, continuous dividend yield, and Richardson extrapolation from 500 and 1000 steps per exercise interval; the two grids agree to $9\times10^{-6}$ (Appendix~\ref{app:tightness}). Decomposing the mean raw gap on these six puts, Ridge LSMC has estimated policy loss $(V^\star{-}\hat L)/K = 0.00108$, with one-sided 95\% upper confidence bound $0.0032$ ($2.1\%$ of mean $V^\star$), and dual looseness $(\hat U{-}V^\star)/K = 0.09622$; all 6 cells satisfy $|V^\star{-}\hat L| \le 1.96\,\mathrm{SE}_L$. Generated programs are comparable: original planner best $0.00216$ (5/6 within noise), score-only rewrites $0.00161$ (6/6). MLP is worse at $0.00722$ (1/6). Dual looseness is identical across programs (the shared $\hat U$ sets it) and ranges from $15\%$ to $236\%$ of $V^\star$ per instance.

These puts also leave little room for improvement. The early-exercise premium, $V^\star$ minus the European value, is the loss of never exercising before maturity. It averages $0.0018$, is below $0.0004$ on four of six contracts, and is smaller than the upper confidence bound on Ridge's own loss. The panel therefore cannot separate Ridge from optimal, or a never-exercise policy from either, and a null search result on it says little about search ability. Since $98.9\%$ of the Ridge gap is dual looseness, this dual can audit any program's value but cannot certify progress at the scale of the search ($0.0015$ against $0.096$). The multi-asset holdout contracts have no binomial reference, so their headroom is unmeasured.

\section{Discussion and limitations}
\label{sec:discussion}

Two questions need separate tools. Whether a candidate improves on its parent is a paired comparison: under a shared $U$ the bound cancels, raw-gap order equals payoff order, and a paired payoff confidence bound on common paths (the public feedback of our original search) answers the question at the resolution of the path budget. A loose certificate does not limit that comparison. The certificate answers whether a candidate is near-optimal: (i)~an estimated cap on suboptimality that holds for any program the agent writes and needs no baseline, and (ii)~with a reference value, a split of that cap into policy loss and bound looseness. Here the split shows that the shared dual is the binding term. A policy-specific dual, such as Andersen--Broadie \citep{andersen2004primal} built from the candidate's own exercise rule, could make the certificate informative at the scale of the search; we did not test it. A benchmark on which search can succeed also needs headroom: contracts with a large early-exercise premium and a positive control, such as a deliberately weakened parent that search should repair. The reliability cost in F3 suggests that how much context a planner reads is itself an experimental variable.

\paragraph{Limitations.}
The study covers three seed blocks, one task family, small search and path budgets, and no positive control; the SD across seeds is descriptive, and path-level error bars are conditional Monte Carlo intervals. The original search consumed the payoff LCB, so its gap curves are post-hoc re-evaluations. The ablation has six matched pairs, fixes history and parent, prescribes both operators, and runs a single round, so memory cannot shape later decisions; its conclusions are limited to this configuration. Reported gaps are Monte Carlo estimates with asymptotic coverage only (\S\ref{sec:cert}), under an evaluator-owned dual that is loose (F5).

\begin{ack}
We thank the four anonymous reviewers, whose comments on the statistical reading of the gap, the role of the shared dual, and the headroom of the reference puts shaped this version.
\end{ack}

\bibliographystyle{plainnat}
\bibliography{refs}

\newpage
\appendix
\setlength{\intextsep}{4pt plus 2pt minus 2pt}

\section{Protocol details}
\label{app:protocol}

\paragraph{Interface.}
A candidate is a complete Python program that exposes \texttt{fit} and \texttt{predict} through a sandboxed CLI. \texttt{fit} receives training paths only; \texttt{predict} is called date by date and may use $S_m$, the date index, and public contract parameters. The evaluator owns path generation under the risk-neutral Black--Scholes dynamics, the exercise loop, the payoff computation, and the dual construction. It never imports or executes candidate code.

\paragraph{Budgets.}
Model: \texttt{gpt-5.6-sol} via a CLI backend; three search seed blocks (named Seed 17, 29 and 41). Public search: $2$ instances $\times 2$ repeats, $256$ training and $512$ pricing paths. Calibration: $2$ instances per seed, $256$ training, $1024$ pricing, $128$ outer $\times 8$ inner. Holdout: $4$ contracts per seed, $512$ training, $2048$ pricing, $256$ outer $\times 16$ inner. Model-call caps: planner $120$\,s, proposer $180$\,s.

\paragraph{Cost.}
Original planner pipeline: $1690.6$\,s of model time ($28.2$~min; $6$ planner and $12$ proposer calls; $\ge 362{,}639$ reported tokens). Direct generation: $1588.7$\,s ($26.5$~min; $\ge 273{,}605$ tokens). Ablation: structured $773.9$\,s ($\ge 183{,}867$ tokens), score-only $812.4$\,s ($\ge 196{,}375$ tokens). Token counts are lower bounds over visible records; minutes are summed model-call durations including timeouts, not wall-clock time.

\paragraph{Reference programs.}
The seven hand-written reference programs are LSMC regressions with different features or models, a two-sweep policy-iteration variant, and a threshold rule. PCA + Ridge and policy iteration give identical results. Both regress on the features $(1, z, z^2, \bar z, Z_m)$ of the log-state $z$; the PCA scores are affine in $z$ and add no new direction, and the second policy-iteration sweep restarts from the terminal cash flows and repeats the first. We report them as one policy. Ridge adds the minimum and maximum of $z$, which coincide with $z$ on one-dimensional contracts, so all three are identical on the puts.

\paragraph{Prediction outcomes.}
Planner pipeline: $3$ refuted, $7$ inconclusive, $2$ execution failures. Direct: $1$ refuted, $8$ inconclusive, $3$ failures. Each generation attempt has a matched control with the same parent, seed, and evaluator budget but no generation.

\section{Per-seed and feedback-pair tables}
\label{app:tables}

\begin{table}[H]
\centering
\small
\caption{Per-seed holdout results. $^\dagger$One policy under two names (Appendix~\ref{app:protocol}).}
\IfFileExists{tables/per_seed.tex}{
\begin{tabular}{l rrr rrr}
\toprule
 & \multicolumn{3}{c}{Gap score $S$ (lower is better)} & \multicolumn{3}{c}{$\hat L_\pi/K$} \\
\cmidrule(lr){2-4}\cmidrule(lr){5-7}
Method / selection rule & Seed 17 & Seed 29 & Seed 41 & Seed 17 & Seed 29 & Seed 41 \\
\midrule
Ridge & 0.07465 & 0.22849 & 0.25267 & 0.17786 & 0.13396 & 0.14132 \\
PCA + Ridge / Policy iteration$^\dagger$ & 0.07490 & 0.22750 & 0.25270 & 0.17760 & 0.13477 & 0.14136 \\
Gated Ridge & 0.07613 & 0.22851 & 0.25341 & 0.17640 & 0.13389 & 0.14073 \\
MLP & 0.08926 & 0.23642 & 0.26112 & 0.16374 & 0.12765 & 0.13372 \\
Ridge + MLP & 0.07607 & 0.22867 & 0.25235 & 0.17674 & 0.13380 & 0.14171 \\
Threshold policy & 0.20579 & 0.32121 & 0.34179 & 0.06665 & 0.05148 & 0.05761 \\
Original planner: best generated & 0.07687 & 0.22908 & 0.25372 & 0.17595 & 0.13337 & 0.14031 \\
Original direct: best generated & 0.07706 & 0.22851 & 0.25355 & 0.17590 & 0.13381 & 0.14069 \\
Original planner incumbent & 0.07687 & 0.22849 & 0.25235 & 0.17595 & 0.13396 & 0.14171 \\
Original direct incumbent & 0.07607 & 0.22851 & 0.25355 & 0.17674 & 0.13381 & 0.14069 \\
Structured-memory incumbent & 0.07465 & 0.22849 & 0.25267 & 0.17786 & 0.13396 & 0.14132 \\
Score-only incumbent & 0.07580 & 0.22849 & 0.25267 & 0.17675 & 0.13396 & 0.14132 \\
\midrule
Common $\hat U/K$ & 0.21423 & 0.26348 & 0.28841 & & & \\
\bottomrule
\end{tabular}
}{(table pending)}
\end{table}

\begin{table}[H]
\centering
\small
\caption{Matched feedback pairs: calibration gap reduction versus the fixed parent (negative is worse; a failed attempt retains the parent and counts as $0$). Deltas in units of $10^{-3}$.}
\IfFileExists{tables/feedback_pairs.tex}{
\begin{tabular}{l l r rr rr}
\toprule
 & & Parent & \multicolumn{2}{c}{Structured memory} & \multicolumn{2}{c}{Score-only memory} \\
\cmidrule(lr){4-5}\cmidrule(lr){6-7}
Seed & Operator & calib.\ $S$ & $\Delta$calib. & $\Delta$holdout & $\Delta$calib. & $\Delta$holdout \\
\midrule
17 & whole-program restart & 0.20115 & failed (0) & --- & -1.648 & -1.616 \\
17 & subsystem rewrite & 0.20115 & -2.533 & +0.180 & +0.303 & -1.156 \\
29 & whole-program restart & 1.86031 & -2.295 & -2.354 & -4.423 & -0.578 \\
29 & subsystem rewrite & 1.86031 & -5.346 & -0.005 & -0.681 & -0.245 \\
41 & whole-program restart & 3.99635 & failed (0) & --- & -2.024 & -0.524 \\
41 & subsystem rewrite & 3.99635 & -2.974 & -0.925 & -0.250 & -0.082 \\
\bottomrule
\end{tabular}
}{(table pending)}
\end{table}

\noindent Raw values, structured / score-only: Seed 17 restart FAILED($0$) / $-0.00165$; Seed 17
rewrite $-0.00253$ / $+0.000303$; Seed 29 restart $-0.00230$ / $-0.00442$; Seed 29 rewrite
$-0.00535$ / $-0.00068$; Seed 41 restart FAILED($0$) / $-0.00202$; Seed 41 rewrite $-0.00297$ /
$-0.00025$.

\section{Holdout comparison figure}
\label{app:figures}

\begin{figure}[H]
\centering
\includegraphics[width=0.8\textwidth]{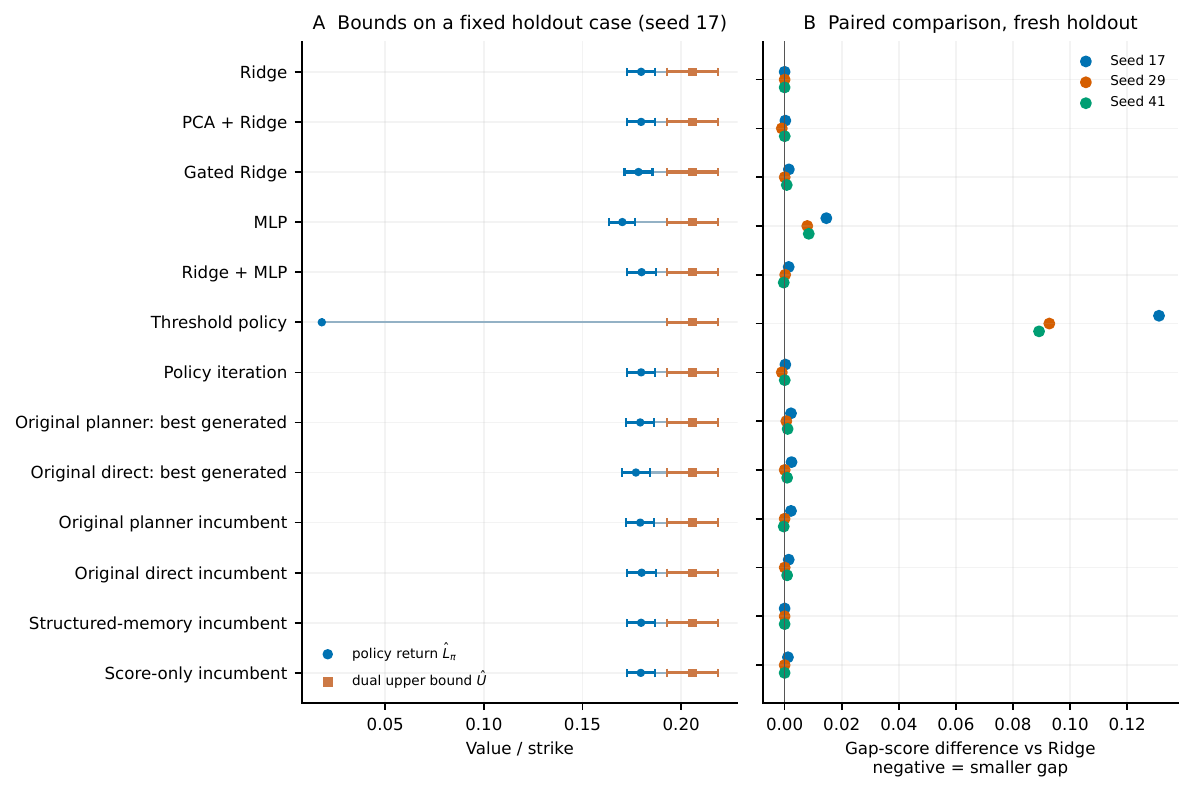}
\caption{Holdout comparison. Left: normalized $\hat L$ and $\hat U$ with $\pm 1.96$\,SE bars for the seed-17 case; these are conditional Monte Carlo intervals on the shared path set. Right: gap-score differences from Ridge across seeds, one point per seed. Ablation incumbents are selected by calibration score alone, with the parent retained as an option.}
\label{fig:results}
\end{figure}

\section{Recovery amendment}
\label{app:recovery}

The first ablation run hit a read-only workspace error while copying the parent program, before any proposer call was issued. Ten already-completed planner plans were continued after a documented recovery: the plans were frozen, byte-identical editable parent copies were created, no planner call was regenerated, and no model call was retried. The two planner timeouts were retained as failures and remain in all denominators. The recovery did not modify the frozen evaluator or any prior result; original records and costs are preserved.

\section{Score decomposition}
\label{app:decomp}

\begin{table}[H]
\centering
\footnotesize
\setlength{\tabcolsep}{4pt}
\caption{Calibration-panel score decomposition; seed-17 rows have $K = 1$.}
\IfFileExists{tables/score_decomposition.tex}{
\begin{tabular}{l l l rrrrrr}
\toprule
Seed & Arm & Operator & $\hat L$ & $\hat U$ & $\hat U-\hat L$ & MC margin & $\tfrac14 Q_{90}$ & $S$ \\
\midrule
17 & parent & Ridge (parent) & 0.19265 & 0.30378 & 0.11113 & 0.04335 & 0.04666 & 0.20115 \\
17 & structured & rewrite & 0.19069 & 0.30378 & 0.11308 & 0.04313 & 0.04747 & 0.20368 \\
17 & score-only & restart & 0.19075 & 0.30378 & 0.11303 & 0.04246 & 0.04731 & 0.20279 \\
17 & score-only & rewrite & 0.19247 & 0.30378 & 0.11131 & 0.04283 & 0.04670 & 0.20084 \\
29 & parent & Ridge (parent) & 0.11862 & 1.25658 & 1.13796 & 0.27673 & 0.44562 & 1.86031 \\
29 & structured & restart & 0.11648 & 1.25658 & 1.14010 & 0.27636 & 0.44614 & 1.86261 \\
29 & structured & rewrite & 0.11429 & 1.25658 & 1.14229 & 0.27609 & 0.44728 & 1.86566 \\
29 & score-only & restart & 0.11439 & 1.25658 & 1.14219 & 0.27563 & 0.44691 & 1.86473 \\
29 & score-only & rewrite & 0.11758 & 1.25658 & 1.13900 & 0.27620 & 0.44579 & 1.86099 \\
41 & parent & Ridge (parent) & 0.12683 & 2.40959 & 2.28275 & 0.48173 & 1.23187 & 3.99635 \\
41 & structured & rewrite & 0.12366 & 2.40959 & 2.28593 & 0.48107 & 1.23232 & 3.99932 \\
41 & score-only & restart & 0.12437 & 2.40959 & 2.28522 & 0.48113 & 1.23203 & 3.99837 \\
41 & score-only & rewrite & 0.12627 & 2.40959 & 2.28332 & 0.48131 & 1.23197 & 3.99660 \\
\bottomrule
\end{tabular}
}{(table pending)}
\end{table}

\noindent Ridge: $\hat L = 0.192648$, $\hat U = 0.303776$, raw gap $0.111128$, mean Monte
Carlo margin $0.043353$, $0.25\,Q_{90} = 0.046665$, $S = 0.201146$. Selected score-only
rewrite: $\hat L = 0.192467$, raw gap $0.111309$, margin $0.042835$,
$0.25\,Q_{90} = 0.046700$, $S = 0.200843$. The score fell only through a smaller sampling
margin while the payoff estimate fell. With two contracts per seed, NumPy's linear
interpolation gives $Q_{90} = 0.1\,g_{\min} + 0.9\,g_{\max}$, hence
$S = 0.525\,g_{\min} + 0.725\,g_{\max}$: unequal contract weighting, not a tail estimate.
The constant $0.25$ is a legacy heuristic predating this study. Because the margin enters $S$ additively, $S$ is not monotone in payoff, and we do not recommend it for selection.

The shared dual bound is loose on some contracts. At seed~41 calibration: $\hat L = 0.082$, $\hat U = 4.614$, $\mathrm{SE}_U = 0.474$, which inflates absolute gaps and shows that the dual estimator is heavy-tailed; candidates cannot tighten the shared $U$.

\section{Certificate tightness on reference-value puts}
\label{app:tightness}

We computed high-precision reference values $V^\star$ for the six one-dimensional Bermudan puts in the holdout panel (two per seed block) using a Cox--Ross--Rubinstein binomial tree. Exercise is allowed only at the evaluator's exercise dates (including $t{=}0$ and mandatory settlement at maturity), with continuous dividend yield and discounting to time zero. We ran 500 and 1000 tree steps per exercise interval and applied Richardson extrapolation; the two grids agree to within $9 \times 10^{-6}$, so $V^\star$ is a numerical reference accurate to about $10^{-5}$, not an exact value. Instance parameters were reconstructed from the frozen evaluator. The European column is the value of never exercising before maturity.

\begin{table}[H]
\centering
\footnotesize
\caption{The six holdout put contracts and their reference values ($K = 1$).}
\label{tab:tightness_instances}
\setlength{\tabcolsep}{3.5pt}
\begin{tabular}{ll rrrrrr r r}
\toprule
Seed & Instance & $S_0$ & $r$ & $q$ & $\sigma$ & $T$ & Dates & European & $V^\star$ \\
\midrule
17 & hidden-000 & 0.9821 & 0.0153 & 0.1016 & 0.2757 & 1.2361 & 7 & 0.17911 & 0.17912 \\
17 & hidden-003 & 0.9196 & 0.0182 & 0.0505 & 0.1880 & 1.5038 & 8 & 0.15741 & 0.15741 \\
29 & hidden-000 & 1.0151 & 0.0659 & 0.0134 & 0.2356 & 1.7781 & 9 & 0.07338 & 0.08247 \\
29 & hidden-003 & 0.9109 & 0.0677 & 0.0818 & 0.2875 & 1.1195 & 5 & 0.16092 & 0.16262 \\
41 & hidden-000 & 0.9477 & 0.0155 & 0.0317 & 0.3174 & 0.9718 & 9 & 0.15429 & 0.15430 \\
41 & hidden-003 & 0.9338 & 0.0646 & 0.1087 & 0.2756 & 1.3263 & 7 & 0.17213 & 0.17243 \\
\bottomrule
\end{tabular}

\end{table}

\begin{table}[H]
\centering
\footnotesize
\caption{Decomposition of the raw certificate width on the holdout puts (means over six cells, $K = 1$). UCB is the one-sided 95\% upper confidence bound $\overline{(V^\star{-}\hat L)} + 1.645\,\bigl(\sum_i \mathrm{SE}_{L,i}^2\bigr)^{1/2}/6$, treating $V^\star$ as exact and cells as independent. ``Within MC noise'' counts cells with $|V^\star{-}\hat L| \le 1.96\,\mathrm{SE}_L$. $^\dagger$Identical on the puts (Appendix~\ref{app:protocol}).}
\label{tab:tightness_methods}
\setlength{\tabcolsep}{4pt}
\begin{tabular}{l r rrrr r}
\toprule
Program & $n$ & $(\hat U-\hat L)/K$ & $(V^\star-\hat L)/K$ & UCB & $(\hat U-V^\star)/K$ & Within MC noise \\
\midrule
Ridge, PCA + Ridge, Pol.\ iter.$^\dagger$ & 6 & 0.09730 & 0.00108 & 0.00320 & 0.09622 & 6/6 \\
Gated Ridge & 6 & 0.09744 & 0.00121 & 0.00333 & 0.09622 & 5/6 \\
MLP & 6 & 0.10344 & 0.00722 & 0.00904 & 0.09622 & 1/6 \\
Ridge + MLP residual & 6 & 0.09779 & 0.00157 & 0.00370 & 0.09622 & 6/6 \\
Threshold policy & 6 & 0.18745 & 0.09123 & 0.09156 & 0.09622 & 0/6 \\
Original planner: best generated & 6 & 0.09838 & 0.00216 & 0.00426 & 0.09622 & 5/6 \\
Original direct: best generated & 6 & 0.09851 & 0.00229 & 0.00435 & 0.09622 & 5/6 \\
Score-only rewrites (one per seed) & 6 & 0.09783 & 0.00161 & 0.00369 & 0.09622 & 6/6 \\
\bottomrule
\end{tabular}

\end{table}

\noindent The calibration is limited to these six puts; the multi-asset holdout instances have no binomial reference. The early-exercise premium ($V^\star$ minus European) is $0.00909$, $0.00169$ and $0.00030$ on three contracts and below $10^{-5}$ on the other three; its mean, $0.0018$, is below every UCB in Table~\ref{tab:tightness_methods}. We fixed no tolerance for near-optimality in advance, so these bounds describe the panel rather than test a hypothesis.

\section{Reproducibility}
\label{app:repro}

The following will be released at \url{https://github.com/MrSteeeve/OpenHyra}: the frozen evaluator and dual construction, the frozen selection manifest, every candidate program including failed attempts, raw evaluation records, the recovery amendment, and the replay scripts. All figures and tables are regenerated from these frozen inputs, and a numerical replay reproduced every reported score and bound. The reference values and the confidence bounds in Appendix~\ref{app:tightness} are computed from the frozen records without new simulation.

\end{document}